\def\year{2022}\relax
\documentclass[letterpaper]{article} 
\usepackage{aaai22}  
\usepackage{times}  
\usepackage{helvet}  
\usepackage{courier}  
\usepackage[hyphens]{url}  
\usepackage{graphicx} 
\usepackage{natbib}  
\usepackage{caption} 
\DeclareCaptionStyle{ruled}{labelfont=normalfont,labelsep=colon,strut=off} 
\usepackage{algorithm}
\usepackage{algorithmic}

\usepackage{booktabs}
\usepackage{amsmath}
\usepackage{multirow}
\usepackage{amssymb}
\usepackage{pifont}
\usepackage{xcolor}
\definecolor{mypink}{rgb}{0.858, 0.188, 0.478}
\usepackage{subfigure}
\usepackage{newfloat}
\usepackage{listings}
\floatstyle{ruled}
\newfloat{listing}{tb}{lst}{}
\floatname{listing}{Listing}
\usepackage{hyperref} 
\hypersetup{
    colorlinks=true,
    linkcolor=red,
}

\title{Transfer Learning from Synthetic to Real LiDAR Point Cloud \\
for Semantic Segmentation}
\author {
    Aoran Xiao, 
    Jiaxing Huang, 
    Dayan Guan, 
    Fangneng Zhan, 
    Shijian Lu\thanks{Corresponding author.}
}
\affiliations {
    
    Nanyang Technological University\\
    aoran.xiao, jiaxing.huang, dayan.guan, fnzhan, shijian.lu@ntu.edu.sg
}

\begin{document}

\maketitle

\begin{abstract}
Knowledge transfer from synthetic to real data has been widely studied to mitigate data annotation constraints in various computer vision tasks such as semantic segmentation. However, the study focused on 2D images and its counterpart in 3D point clouds segmentation lags far behind due to the lack of large-scale synthetic datasets and effective transfer methods. We address this issue by collecting SynLiDAR, a large-scale synthetic LiDAR dataset that contains point-wise annotated point clouds with accurate geometric shapes and comprehensive semantic classes. SynLiDAR was collected from multiple virtual environments with rich scenes and layouts which consists of over 19 billion points of 32 semantic classes.
In addition, we design PCT, a novel point cloud translator that effectively mitigates the gap between synthetic and real point clouds. Specifically, we decompose the synthetic-to-real gap into an appearance component and a sparsity component and handle them separately which improves the point cloud translation greatly. We conducted extensive experiments over three transfer learning setups including data augmentation, semi-supervised domain adaptation and unsupervised domain adaptation. Extensive experiments show that SynLiDAR provides a high-quality data source for studying 3D transfer and the proposed PCT achieves superior point cloud translation consistently across the three setups. SynLiDAR project page: \url{https://github.com/xiaoaoran/SynLiDAR}
\end{abstract}

\begin{figure*}[t]
    \centering
    \includegraphics[width=\textwidth]{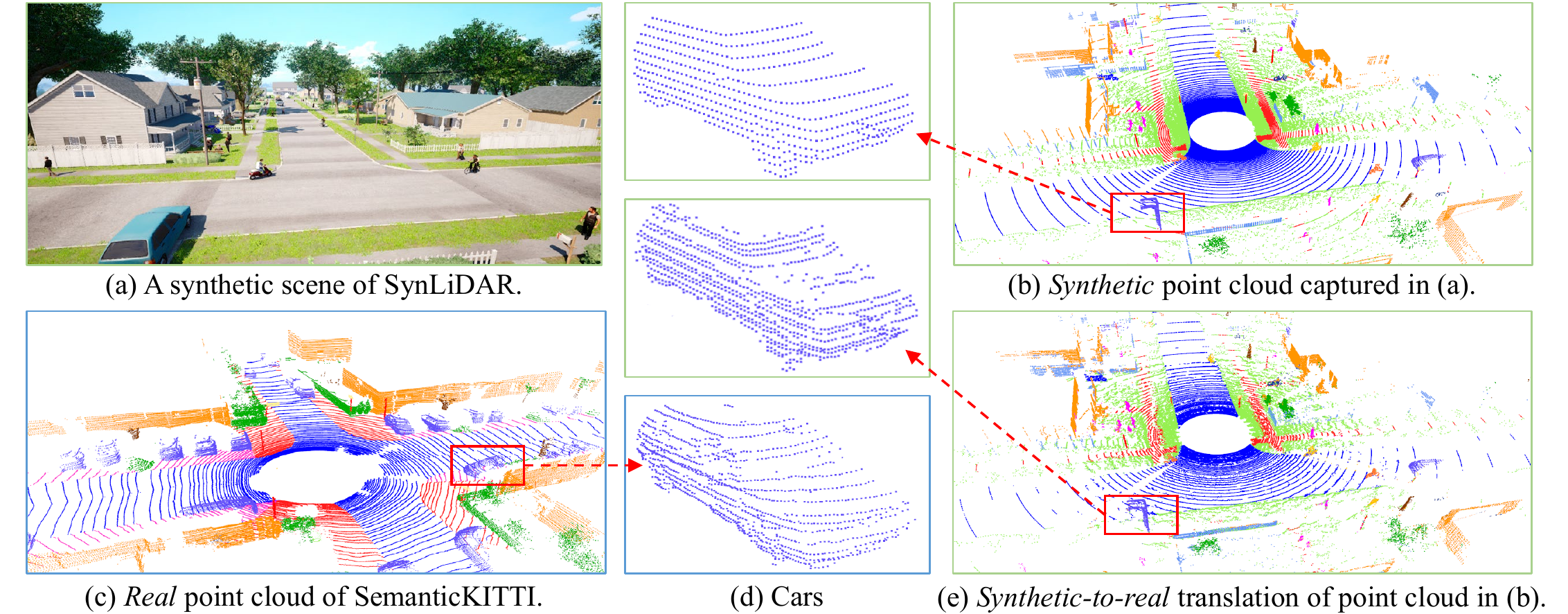}
    \captionof{figure}{
    We create SynLiDAR, a large-scale multiple-class synthetic LiDAR point cloud dataset as illustrated in (b). SynLiDAR contains over 19 billion annotated points of 32 semantic classes which was collected by constructing multiple virtual environments and 3D object models as shown in (a). To make synthetic point cloud more useful for handling real-world LiDAR point cloud as shown in (c), we design a point cloud translator (PCT) that translates synthetic point cloud by decomposing the domain gap into an appearance component and a sparsity component. The translated data in (e) has a closer distribution as real point cloud and is more effective in processing real point cloud. The close-up views in (d) show the translation effects.
    }
    \label{fig.example}
\end{figure*}

\section{Introduction}
Semantic segmentation of LiDAR sequential point cloud is critical in various scene perception tasks and it has attracted increasing attention from both industry and academia \citep{behley2019semantickitti,milioto2019rangenet++,pan2020semanticposs,hu2020randla,tang2020searching} in recent years. However, training effective segmentation models requires large amounts of annotated point cloud, which are prohibitively time-consuming to collect due to the view change of 3D data and visual inconsistency between LiDAR point cloud and physical world. This can be observed by the small size of existing \textit{real} LiDAR sequential datasets as listed in Table~\ref{tab.Compare_datasets}.

Inspired by the great success of 
transfer learning from synthetic to real data
in two-dimensional (2D) field \citep{ros2016synthia, armeni20163d,gaidon2016virtual}, one possible way to mitigate the data annotation constraint is to leverage synthetic point cloud data that can be collected and annotated automatically by computer engines. However, collecting large-scale synthetic LiDAR sequential point cloud is a nontrivial task which involves a large number of virtual scenes and objects as well as complicated point generation processes. In addition, most existing transfer learning methods \citep{sun2019not,saito2018maximum,kodirov2015unsupervised,saito2019semi,wu2019squeezesegv2} focus on 2D images which do not work for 3D point cloud. To the best of our knowledge, few researches tackle the challenge of synthetic-to-real transfer of point cloud of nature scenes, largely due to the lack of large-scale synthetic data with accurate geometries and rich semantic annotations. 

We address the said issues by creating SynLiDAR, a large-scale LiDAR sequential point cloud dataset for facilitating the research of synthetic-to-real transfer of 3D point cloud data. We collected SynLiDAR from multiple virtual environments that were constructed by professional 3D generalists with advanced graphic tools. Each virtual environment contains configurable object models that are similar to real-world data in both geometry and layout. The dataset is ideal for the study of synthetic-to-real transfer as it consists of comprehensive and diverse point semantics (32 semantic classes with over 19 billion point-wise annotated points) and its collected points are also highly accurate in geometry.

In addition, we designed PCT, a point-cloud translator for mitigating domain gaps between synthetic and real point cloud as illustrated in Fig. \ref{fig.example}. The design of PCT is inspired by the observations that point clouds can be viewed as discrete samplings of continuous 3D geometric environments where the domain gaps between synthetic and real point clouds come from either appearance differences (due to environments variations) or sparsity differences (due to sampling variation by sensors). We hence disentangle the  domain gap into an appearance component and a sparsity component, and design an appearance translation module (ATM) and a sparsity translation module (STM) to handle the two gap components separately. Specifically, ATM first up-samples synthetic point cloud and translates it to have similar appearance as real point cloud. STM then extracts sparsity features from real point cloud and fuses it with the ATM output to translate synthetic point cloud to have real sparsity. To the best of our knowledge, PCT is the first translation method for LiDAR point clouds in natural scenes.

The contribution of this work can be summarized in three aspects as listed:
\begin{itemize}
    \item We create SynLiDAR, a large-scale synthetic LiDAR sequential point cloud dataset that has rich semantic classes and a large number of points with accurate point-wise annotations. SynLiDAR will lay a strong foundation for the study of the under-explored synthetic-to-real transfer in LiDAR point cloud segmentation.
    
    \item We examine the major underlying factors of the domain gap between synthetic and real point clouds, and design PCT, a pioneer LiDAR point cloud translator that can transform synthetic point clouds to have similar features and distributions as real point clouds and accordingly mitigate the domain gap effectively.
    \item We design three experimental setups for the study of synthetic-to-real point cloud transfer: data augmentation (DA), semi-supervised domain adaptation (SSDA) and unsupervised domain adaptations (UDA). We conducted extensive experiments under the three setups which will form valuable bases for the future investigation of synthetic-to-real point cloud transfer.
\end{itemize}

\setlength{\tabcolsep}{2.5mm}{
\begin{table*}[t]
  \centering
  \begin{tabular}{lcrrccccc}
    \toprule
    dataset & format & \#scans & \#points & \#classes & annotation & type \\
    \midrule
    SemanticKITTI \citep{behley2019semantickitti} & point & 43,552 & 4,549M & 25 & point-wise & real\\
    SemanticPOSS \citep{pan2020semanticposs} & point & 2,988 & 216M  & 14 & point-wise & real\\
    nuScenes-Lidarseg \citep{caesar2020nuscenes} & point & 40,000 & 1,400M & 32 & point-wise & real\\
    GTA-LiDAR \citep{wu2018squeezeseg}& image  & 121,087 & - & 2 & pixel-wise & synthetic\\
    PreSIL \citep{hurl2019precise} & point & 51,074 & 3,135M & 2 & point-wise & synthetic\\
    SynLiDAR (ours)& point  & 198,396 & 19,482M & 32 & point-wise & synthetic\\
    \bottomrule
 \end{tabular}
\caption{Overview of outdoor LiDAR sequential point cloud datasets with semantic annotations: \#scans: Number of scans for the datasets; \#points: Number of points in millions (M); \#classes: Number of semantic classes.}
 \label{tab.Compare_datasets}
\end{table*}}

\section{Related Works}

\subsection{Semantic Segmentation of Point Cloud}
3D deep learning \citep{guo2020deep} has attracted increasing attention, which is important for different LiDAR perception tasks including semantic segmentation.
Different approaches have been proposed to segment LiDAR point clouds, including projection-based methods ~\citep{wu2018squeezeseg,wu2019squeezesegv2,milioto2019rangenet++,xiao2021fps}, point-based methods~\citep{qi2017pointnet, qi2017pointnet++, hu2020randla}, sparse convolutional methods~\citep{graham20183d, choy20194d}, customized 3D convolutional methods~\citep{thomas2019kpconv,zhu2020cylindrical}, etc.

\subsection{LiDAR Sequential Point Cloud Datasets}
 LiDAR \textit{sequential} point clouds provide point cloud scans, each containing sparse and incomplete points collected in a sweep by LiDAR sensors. Several real world datasets, including SemanticKITTI~\citep{behley2019semantickitti}, SemanticPOSS~\citep{pan2020semanticposs} and nuScenes-Lidarse~\citep{caesar2020nuscenes}, have been proposed recently and promote the developments of LiDAR point cloud segmentation researches.
 However, labeling point-wise semantic annotations is prohibitively time-consuming for LiDAR sequential point clouds. Therefore, existing real point cloud datasets have very limited data sizes as listed in Table \ref{tab.Compare_datasets}.

Inspired by the success of 2D synthetic image datasets \citep{richter2016playing}, a few pioneer studies \citep{wu2018squeezeseg,hurl2019precise} have explored to collect synthetic point cloud data from GTA-V games. However, 3D meshes in GTA-V games are not accurate
and GTA-V games provide only two object classes \textit{Car} and \textit{Pedestrian}~\citep{wu2018squeezeseg}. Its collected synthetic data are thus insufficient for studying fine-grained LiDAR point cloud segmentation. We instead construct a wide range of realistic virtual environments and object models by leveraging graphic tools and professionals. The synthetic point clouds within the SynLiDAR thus capture much more accurate geometries and the rich diversity of semantic labels as in natural scenes.

\subsection{Transfer Learning of Point Cloud}
Transfer learning aims to transfer the knowledge from the source domain to the target domain. It is an important tool to solve the inefficient training data problem ~\citep{tan2018survey}. This paper discusses three important transfer learning tasks of point cloud: DA combines multiple labeled datasets for training to reach better performances than training on each single one~\citep{li2020pointaugment,chen2020pointmixup}; SSDA exploits the knowledge from the source data with annotations and use a certain number of unlabeled examples and a few labeled ones from the target domain to learn a target model; UDA instead uses annotated source data and target data without annotations to learn the target model~\citep{qin2019pointdan,saleh2019domain,yang2021st3d}.
Several pioneer works \citep{wu2019squeezesegv2,zhao2020epointda,yi2021complete} have been proposed for the research of UDA problem in the LiDAR segmentation task.
Instead, PCT mitigates domain gap problem in the input space and is effective for different kinds of transfer learning setups.

\subsection{Domain Translation of Point Cloud}
Domain Translation aims to learn meaningful mapping across domains.
It is well developed for 2D images between paired domains~\citep{isola2017image}, unpaired domains~\citep{zhu2017cycle}, multiple modalities~\citep{zhu2017toward}, etc. For 3D data, some attempt has been reported for translation from images to depth ~\citep{liu2015deep}, from point cloud to depth \citep{roveri2018network}, from point cloud to images \citep{dai2020neural}, etc. However, existing generative methods~\citep{li2019pugan,xie2021sparenet} mainly focus on 3D objects while the translation between LiDAR point clouds in scenes is largely neglected. We address this challenge for mitigating the gap between synthetic and real point clouds. 

\section{The SynLiDAR Dataset} \label{Sec.synlidar}
SynLiDAR is collected from multiple realistic virtual scenes constructed by professional 3D generalists using the Unreal Engine 4 platform \citep{UE4Market} (as shown in Fig. \ref{fig.example} a).
These virtual scenes include different types of outdoor environments such as cities, towns, harbour, etc, covering large area of virtual areas.
They are constituted by a large number of physically accurate object models that are produced by expert modelers with 3D-Max software, to ensure the 
high quality of synthetic data.
Specifically, accurate coordinates and precise point-wise annotations of point cloud are collected automatically from these virtual environments.
Note that SynLiDAR can be easily expanded by including new virtual scenes, 3D objects of new classes, etc. More descriptions about our constructed virtual scenes and data collection process could be found in Appendix.

\begin{figure}
    \centering
    \includegraphics[width=0.5\textwidth]{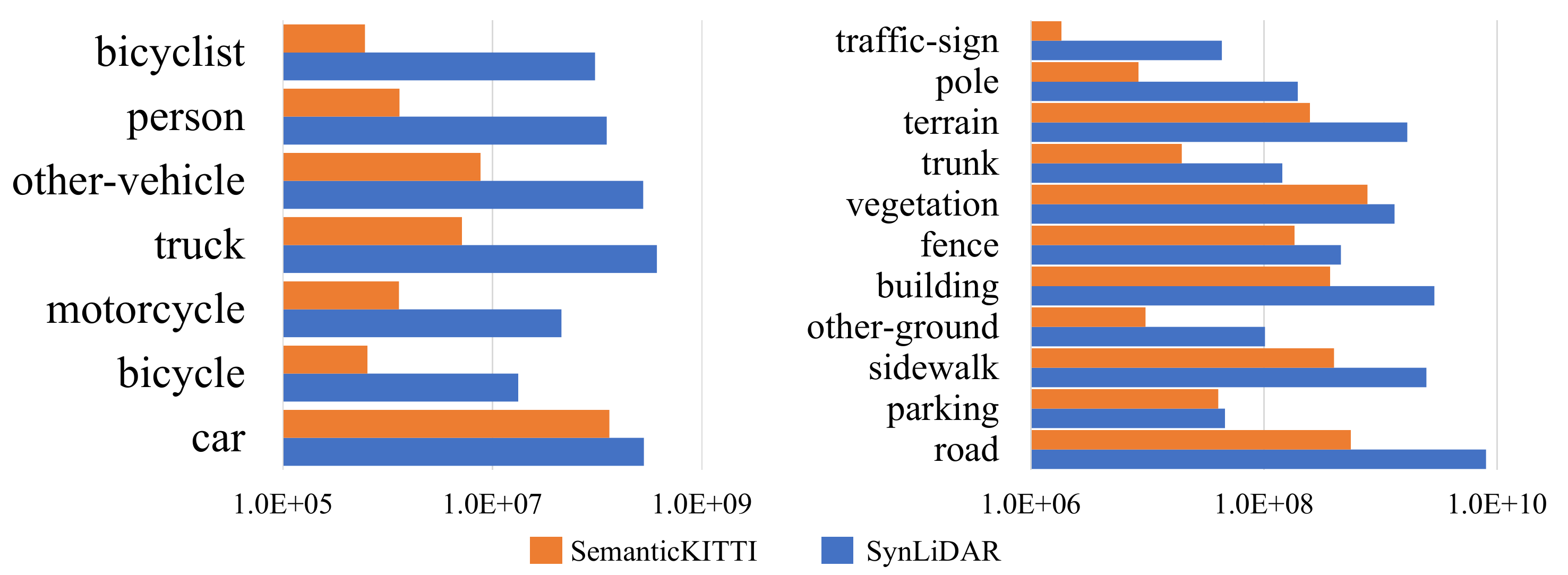}
    \caption{The numbers of annotated points (x-axis) per class (y-axis) for SemanticKITTI and SynLiDAR. Left: \textit{Thing} classes; Right:\textit{Stuff} classes.}
    \label{fig:compare_dataset}
\end{figure}

Another important attribute of LiDAR point cloud is \textit{intensity}, which is challenging to simulate due to the complicated signal transmission, propagation and reception processes in real environments \cite{wu2019squeezesegv2}. 
In SynLiDAR, we address this issue by training a rendering model that learns from real LiDAR point cloud and predicts intensity values for SynLiDAR. 
Although there are no ground truth of intensity values in SynLiDAR, we proved the effectiveness of our predicted values through data augmentation experiments.
Detailed descriptions about the rendering model and experiments are presented in Appendix.


\begin{figure*}[t]
  \centering
  \includegraphics[width=\textwidth]{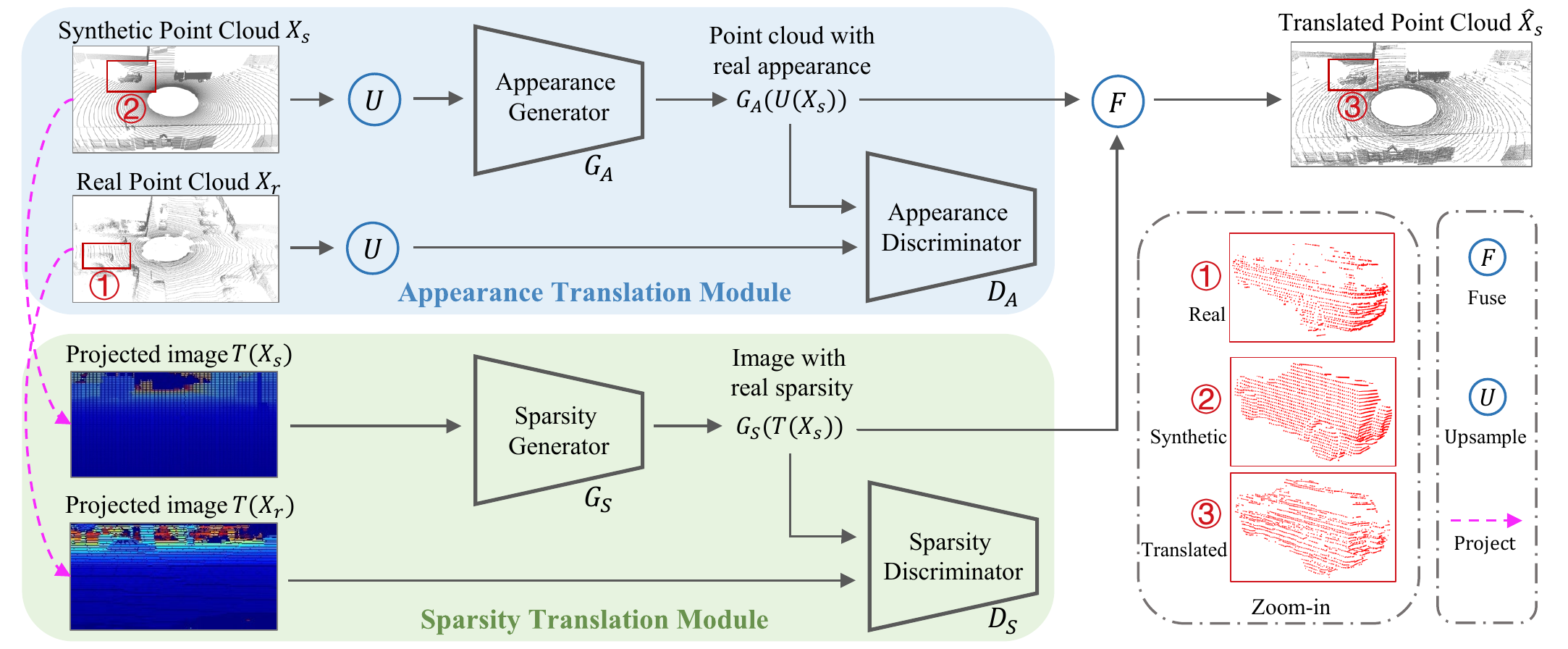}
  \caption{
  The proposed PCT disentangles point-cloud translation into appearance translation and sparsity translation tasks. Given synthetic point cloud, the appearance translation first learns to reconstruct dense point cloud that have similar appearance as real point cloud. The sparsity translation then learns real sparsity distribution in 2D space and fuses it with the reconstructed point cloud in 3D space. The final translation has similar appearance and sparsity as real point cloud as illustrated.
} 
  \label{fig.tech_pipeline}
\end{figure*}

SynLiDAR has 13 LiDAR point cloud sequences with point-wise annotations.
It has 198,396 scans of point cloud with 19 billion points in total, where each scan has around 98,000 points on average. Precise point-wise annotations of 32 semantic classes are provided for fine-grained scene understanding.
Tab. \ref{tab.Compare_datasets} shows that SynLiDAR outweighs the existing LiDAR point cloud dataset of semantic segmentation in both point numbers and semantic classes and
Fig. \ref{fig:compare_dataset} compares point numbers for categories of \textit{thing} (countable foreground classes) and  \textit{stuff} (uncountable background classes) in SynLiDAR and SemanticKITTI (the largest real point cloud dataset to the best of our knowledge), indicating that SynLiDAR is a truly \textit{large-scale} point cloud dataset and an ideal data source for transfer learning researches on synthetic-to-real point cloud.

\section{Point Cloud Translation}

Similar to most synthetic data, point clouds in SynLiDAR have a clear domain gap with real LiDAR data, and the model trained using SynLiDAR usually experiences clear performance drops while applied to real point clouds. 
This paper proposes PCT, the first translator of LiDAR data to mitigate the domain gap for the challenging scene semantic segmentation task, as illustrated in this section.

An intuitive idea to mitigate the domain gap is to employ existing 3D generative models~\citep{achlioptas2018learning,li2019pugan,huang2020pf} to translate synthetic data to have real data distributions. However, these models are designed for 3D objects with uniformly distributed points, and standard supervisions like Chamfer loss or Earth Mover's Distance (EMD) \citep{fan2017point} fail to regularize LiDAR data distribution of 3D scenes directly. As a result, few studies have attempted to address the problem of LiDAR point cloud translation.

We observed that point clouds are discrete samplings of the continuous geometric world, hence the synthetic-to-real gaps could be disentangled into two components: The appearance component reflects the differences between synthetic and real continuous environments, and the sparsity component shows differences in sampling patterns introduced by different LiDAR sensors.
The proposed \textit{PCT} mitigates these two components separately and its translated synthetic point clouds have both similar appearance and sparsity as targeting real-world data.

\renewcommand\arraystretch{1.0}
\setlength{\tabcolsep}{0.7mm}{
\begin{table*}[t]
\centering
\small
\begin{footnotesize}
\begin{tabular}{c|ccccccccccccccccccc|c}
 \toprule
  method & \rotatebox{90}{car} & \rotatebox{90}{bi.cle} & \rotatebox{90}{mt.cle} & \rotatebox{90}{truck} & \rotatebox{90}{oth-v.} & \rotatebox{90}{pers.} & \rotatebox{90}{bi.clst} & \rotatebox{90}{mt.clst} & \rotatebox{90}{road} & \rotatebox{90}{parki.} & \rotatebox{90}{sidew.} & \rotatebox{90}{oth-g.} & \rotatebox{90}{build.} & \rotatebox{90}{fence} & \rotatebox{90}{veget.} & \rotatebox{90}{trunk} & \rotatebox{90}{terra.} & \rotatebox{90}{pole} & \rotatebox{90}{traf.} & mIoU\\
  \midrule
  baseline & 95.7 & 25.0 & 57.0 & 62.1 & 46.4 & 63.4 & 77.3 & 0.0 & 93.0 & 47.9 & 80.5 & 2.2 & 89.7 & 58.6 & 89.5 & 66.5 & 78.0 & 64.6 & 50.1  & 60.3\\
   Jittering~\citep{qi2017pointnet} & 95.7 & 27.8 & 56.2 & 66.0 & 45.8 & 65.3 & 82.8 & 0.0 & 93.0 & 48.2 & 79.9 & 2.5 & 89.7 & 62.9 & 88.9 & 64.0 & 77.0 & 64.8 & 51.0 & 61.2 \\
   Dropout~\citep{srivastava2014dropout} & 96.0 & 28.5 & 57.1 & 65.1 & 46.4 & 64.1 & 83.6 & 0.1 & 93.5 & 47.6 & 80.1 & 2.3 & 89.3 & 61.9 & 90.1 & 66.9 & 78.8 & 65.8 & 49.1 & 61.4 \\
   PointAug~\citep{li2020pointaugment} & 95.9 & 29.2 & 70.0 & 76.3 & 50.0 & 67.0 & 84.4 & 2.4 & 93.8 & 48.1 & 81.2 & 4.6 & 89.8 & 58.4 & 87.5 & 65.4 & 72.7 & 62.4 & 50.5 & 62.6 \\
   \midrule
  +SynLiDAR & 95.9 & 33.0 & 62.8 & 78.9 & 50.2 & 71.4 & 83.5 & 0.7 & 92.3 & 52.8 & 79.9 & 0.1 & 89.8 & 59.5 & 86.3 & 65.4 & 72.8 & 63.6 & 48.9 & 62.5 \\
   PCT & 96.3 & 38.7 & 73.4 & 82.9 & 56.1 & 71.1 & 85.3 & 1.6 & 94.1 & 54.3 & 81.6 & 1.3 & 89.5 & 59.6 & 87.8 & 66.9 & 73.6 & 65.4 & 50.5 & 64.7 \\
 \bottomrule
\end{tabular}
\end{footnotesize}
\caption{
Data Augmentation experiments on SemanticKITTI:
Combining the training data of SynLiDAR and SemanticKITTI trains more accurate semantic segmentation models. PCT mitigates the domain gap effectively and combining the PCT-translated SynLiDAR with SemanticKITTI further improves the segmentation.
}
\label{tab:da_kitti}
\end{table*}}

\renewcommand\arraystretch{1.0}
\setlength{\tabcolsep}{1.5mm}{
\begin{table*}[t]
\centering
\begin{footnotesize}
\begin{tabular}{c|ccccccccccccc|c}
 \toprule
method & pers. & rider & car & trunk & plants & traf. & pole & garb. & buil. & cone. & fence & bike & grou. & mIoU\\
   \midrule
   baseline & 55.6 & 45.1 & 66.9 & 44.4 & 73.9 & 45.4 & 41.6 & 14.5 & 76.1 & 7.9 & 57.0 & 54.1 & 75.3 & 50.6 \\
   Jittering~\citep{qi2017pointnet} & 55.2 & 48.7 & 65.1 & 45.5 & 75.2 & 45.9 & 40.9 & 18.1 & 76.4 & 15.1 & 57.1 & 56.4 & 75.0 & 51.9 \\
   Dropout~\citep{srivastava2014dropout} & 56.9 & 56.7 & 67.8 & 43.3 & 75.6 & 40.3 & 30.7 & 26.8 & 75.7 & 17.7 & 57.3 & 55.6 & 78.6 & 52.5 \\
   PointAug~\citep{li2020pointaugment} & 62.3 & 60.7 & 69.6 & 39.3 & 76.0 & 41.4 & 33.8 & 24.1 & 78.0 & 13.7 & 62.2 & 56.5 & 79.2 & 53.6 \\
   \midrule
   +SynLiDAR & 57.6 & 59.3 & 61.1 & 37.1 & 76.1 & 32.7 & 40.9 & 34.7 & 72.7 & 37.7 & 57.6 & 43.3 & 81.2 & 53.2 \\
   PCT & 57.3 & 61.4 & 65.8 & 36.2 & 77.4 & 42.5 & 42.1 & 49.2 & 74.5 & 32.4 & 55.8 & 48.9 & 81.8 & 55.8 \\
 \bottomrule
\end{tabular}
\end{footnotesize}
\caption{
Data Augmentation experiments on SemanticPOSS:
Combining the training data of SynLiDAR and SemanticPOSS trains more accurate semantic segmentation models. PCT mitigates the domain gap effectively and combining the PCT-translated SynLiDAR with SemanticPOSS further improves the segmentation.
}
\label{tab:da_poss}
\end{table*}}

\textbf{The pipeline of PCT}: Given synthetic point clouds $X_s\in \mathbb{R}^{N_s\times 3}$ and real point clouds $X_r\in \mathbb{R}^{N_r\times 3}$, we aim to translate $X_s$ into $\hat{X}_{s\rightarrow r}\in \mathbb{R}^{\hat{N}_s\times 3}$ that have similar appearance and sparsity as $X_r$. 
Firstly, the ATM up-samples $X_s$ into $U(X_s)$, and translates it to have certain real appearance by a generator $G_A$ as $X'_s=G_A(U(X_s))$. 
Then the STM projects ($T$) point cloud into images and extracts sparsity features of real point cloud into synthetic data by another generator $G_S$, \textit{i.e.} $X''_s=G_S(T(X_s))$. Finally, translated point cloud is generated by fusing $X'_s$ and $X''_s$ as $\hat{X}_{s\rightarrow r}=F(X'_s, X''_s)$.
More details of the two translation modules are provided in the ensuing two subsections.

\textbf{ATM} aims to translate synthetic point cloud to have real appearance and we realize it through a 3D generative adversarial network that consists of a generator $G_A$ and a discriminator $D_A$, and train them in an adversarial manner. In this stage, we first up-sample both synthetic and real point cloud so as to eliminate the effect of domain-specific sparsity features. The generator aims to produce intermediate representation $X'_s$ (from the up-sampled synthetic data) to have real appearance to fool the discriminator, while the discriminator learns to distinguish $X'_s$ from $U(X_r)$. Specifically, The adversarial learning loss can be formulated as follows:
\begin{equation}
\begin{aligned}
    L^{adv}_{A} = & \log(D_A(-G_A(U(X_s))\log(G_A(U(X_s))))) +  \\
    & \log(1-D_A(U(X_r)\log(U(X_r))))
\end{aligned}
\end{equation}
We introduce EMD for generator to keep geometries of $X_s$ and $X'_s$
\begin{equation}
    L^{emd}_{A}(X_s, X'_s)=\sum\limits_{x\in X_s}{||x-\phi(x)||_2}
\end{equation}
where $\phi:X_s\rightarrow X'_s$ is a bijection.
The objective function of the ATM can be formulated as:
\begin{equation}
\begin{aligned}
    L_A(G_A,D_A)=\arg\min\limits_{G_A}\max\limits_{D_A}( & \lambda^{adv}_{A}L^{adv}_{A} + \\ 
    & \lambda^{emd}_{A}L^{emd}_{A})
\end{aligned}
\end{equation}

\textbf{STM} aims to transfer sparsity information from real point cloud to synthetic point cloud. However, existing supervisions \citep{fan2017point} such as Chamfer loss and EMD loss cannot capture sparsity information well as they lack sparsity regulation. To address this problem, we propose to first learn sparsity information in 2D space and then fuse it back into 3D space. Specifically, we first project $X_s$ and $X_r$ into depth images (as $T(X_s)$ and $T(X_r)$ respectively), and then employ an image-to-image translation model to translate $T(X_s)$ to have similar sparsity as $T(X_r)$. The translation network is also GAN-based with a generator $G_S$ and a discriminator $D_S$. The adversarial learning loss can be formulated as follows
\begin{equation}
\begin{aligned}
    L^{adv}_{S} & = \log(D_S(-G_S(T(X_s))\log(G_S(T(X_s))))) \\
    & + \log(1-D_S(T(X_r)\log(T(X_r))))
\end{aligned}
\end{equation}
To preserve the geometry information during the translation, we further include a geometry consistency loss to ensure that the translated depth image preserves similar geometry as the original image:
\begin{equation}
\begin{aligned}
    L^{geo}_{S} & = ||\overline{T(X_s)}-\overline{G_S(T(X_s))}||_2 \\
    & + ||\overline{T(X_r)}-\overline{G_S(T(X_r))}||_2
\end{aligned}
\end{equation}
where $\overline{A}-\overline{B}$ means that the distance is computed for pixels existing in both A and B only. 
The overall objective of the sparsity translation module can be formulated by:
\begin{equation}
\begin{aligned}
    L_S(G_S,D_S)=\arg\min\limits_{G_S}\max\limits_{D_S}( & \lambda^{adv}_{S}L^{adv}_{S} + \\ 
    & \lambda^{geo}_{S}L^{geo}_{S})
\end{aligned}
\end{equation}

Finally, the translated image $G_S(T(X_s))$ with real sparsity information are projected back into 3D space for fusion. Specifically, $G_S(T(X_s))$ are used for guidance to drop out points in $X'_s$. The semantic labels of the translated point cloud are assigned according to labels of neighboring points in the original point cloud.

\renewcommand\arraystretch{1.0}
\setlength{\tabcolsep}{0.8mm}{
\begin{table*}[t]
\centering
\begin{footnotesize}
\begin{tabular}{c|ccccccccccccccccccc|c}
 \toprule
  Method & \rotatebox{90}{car} & \rotatebox{90}{bi.cle} & \rotatebox{90}{mt.cle} & \rotatebox{90}{truck} & \rotatebox{90}{oth-v.} & \rotatebox{90}{pers.} & \rotatebox{90}{bi.clst} & \rotatebox{90}{mt.clst} & \rotatebox{90}{road} & \rotatebox{90}{parki.} & \rotatebox{90}{sidew.} & \rotatebox{90}{oth-g.} & \rotatebox{90}{build.} & \rotatebox{90}{fence} & \rotatebox{90}{veget.} & \rotatebox{90}{trunk} & \rotatebox{90}{terra.} & \rotatebox{90}{pole} & \rotatebox{90}{traf.}  & mIoU \\
 \midrule
  S+T & 56.2 & 3.0 & 15.1 & 1.0 & 5.0 & 20.2 & 42.1 & 2.8 & 52.1 & 0.7 & 19.8 & 0.0 & 41.3 & 5.8 & 62.1 & 34.0 & 42.0 & 24.6 & 1.4 & 22.6 \\
   MMD~\citep{tzeng2014mmd} & 56.4 & 3.3 & 13.3 & 1.5 & 6.1 & 21.4 & 34.6 & 1.6 & 54.3 & 0.4 & 21.4 & 0.0 & 50.2 & 5.8 & 61.2 & 37.0 & 44.9 & 31.6 & 2.2 & 23.5 \\
   MME~\citep{saito2019mme} & 51.0 & 5.6 & 13.1 & 1.3 & 7.3 & 15.1 & 54.4 & 4.4 & 43.1 & 0.2 & 28.3 & 0.0 & 60.7 & 13.3 & 66.1 & 30.1 & 39.9 & 24.8 & 6.6 & 24.5 \\
   APE~\citep{kim2020APE} & 58.6 & 6.2 & 16.6 & 3.1 & 11.3 & 14.2 & 35.8 & 3.7 & 61.5 & 1.7 & 30.3 & 0.0 & 54.7 & 15.4 & 64.6 & 20.0 & 45.5 & 23.9 & 9.1 & 25.1 \\
  \midrule
   PCT & 56.0 & 7.0 & 17.1 & 2.8 & 9.9 & 23.7 & 43.7 & 5.6 & 55.3 & 0.8 & 22.9 & 0.0 & 50.1 & 8.4 & 65.3 & 23.1 & 43.5 & 28.8 & 7.5 & 24.8 \\
   APE + PCT & 58.1 & 7.3 & 17.8 & 2.6 & 13.9 & 24.7 & 46.5 & 5.1 & 60.5 & 1.9 & 31.3 & 0.0 & 56.8 & 14.6 & 67.9 & 23.7 & 44.3 & 26.1 & 9.3 & 27.0 \\
 \bottomrule
\end{tabular}
\end{footnotesize}
\caption{
Experiments on semi-supervised domain adaptation with SynLiDAR (as source) and SemanticKITTI (as target): PCT translates SynLiDAR and mitigates domain gaps in the input space effectively. It is complementary to APE and combining them outperforms the baseline SynLiDAR + SemanticKITTI (\textit{i.e.}, S+T) by large margins. 
}
\label{tab:ssda_kitti}
\end{table*}}

\renewcommand\arraystretch{1.0}
\setlength{\tabcolsep}{1.8mm}{
\begin{table*}[t]
\centering
\begin{footnotesize}
\begin{tabular}{c|ccccccccccccc|c}
 \toprule
  Method & pers. & rider & car & trunk & plants & traf. & pole & garb. & buil. & cone. & fence & bike & grou. & mIoU\\
 \midrule
   S+T & 25.2 & 36.1 & 18.2 & 12.8 & 58.6 & 1.7 & 30.5 & 5.6 & 25.7 & 3.0 & 12.0 & 10.6 & 75.6 & 24.3 \\
   MMD~\citep{tzeng2014mmd} & 25.5 & 35.7 & 28.9 & 6.7 & 64.3 & 1.7 & 23.2 & 5.6 & 53.3 & 3.3 & 30.2 & 13.9 & 70.4 & 27.9 \\
   MME~\citep{saito2019mme} & 33.2 & 40.2 & 25.0 & 11.0 & 61.9 & 0.4 & 31.2 & 7.3 & 56.1 & 5.7 & 37.1 & 6.7 & 71.2 & 29.8 \\
   APE~\citep{kim2020APE} & 34.3 & 40.1 & 21.5 & 16.3 & 62.6 & 0.9 & 31.1 & 2.3 & 55.9 & 13.3 & 34.3 & 9.6 & 71.6 & 30.3 \\
   \midrule
   PCT & 25.8 & 36.8 & 27.8 & 11.3 & 62.2 & 1.9 & 31.2 & 5.2 & 58.7 & 2.6 & 34.3 & 8.5 & 68.7 & 28.8 \\
   APE + PCT & 34.7 & 36.3 & 27.2 & 15.8 & 62.9 & 0.8 & 31.6 & 8.7 & 62.3 & 9.8 & 35.1 & 9.3 & 70.9 & 31.2 \\
 \bottomrule
\end{tabular}
\end{footnotesize}
\caption{
Experiments on semi-supervised domain adaptation with SynLiDAR (as source) and SemanticPOSS (as target): PCT translates SynLiDAR and mitigates domain gaps in the input space effectively. It is complementary to APE and combining them outperforms the baseline SynLiDAR + SemanticPOSS (\textit{i.e.}, S+T) by large margins. 
}
\label{tab:ssda_poss}
\end{table*}}

\section{Experiments}

We evaluate how SynLiDAR benefits semantic segmentation over multiple real point cloud datasets and how PCT mitigates domain gaps between SynLiDAR and real point cloud data. 
We conducted experiments on three transfer setups including DA, SSDA, and UDA as introduced in section 2.3. All experiments were conducted by using state-of-the-art 3D semantic segmentation network MinkowskiNet \citep{choy20194mink}.

\subsection{Datasets and Implementation Details}
We conduct experiments over two real-world point cloud datasets. The first is \textit{SemanticKITTI} \citep{behley2019semantickitti} that consists of 43,552 scans of annotated sequential LiDAR point cloud with 19 semantic classes. It is the largest real-world sequential LiDAR point cloud dataset for semantic segmentation to the best of our knowledge. We follow the commonly-used protocol that splits sequences 00-07, 09-10 for training and sequence 08 for validation. The second is \textit{SemanticPOSS} \citep{pan2020semanticposs} which is collected in a university campus. It consists of 2,988 annotated point cloud scans of 14 semantic classes. We follow the benchmark setting by using sequence 03 for validation and the rest for training. 
We \textit{ignore} extra classes of SynLiDAR by mapping them as 'unlabeled' for each real dataset.
Mean Intersection of Union (mIoU) is used as the evaluation metric.

For point cloud translation, parameters $\lambda^{adv}_{A}$, $\lambda^{emd}_{A}$, $\lambda^{adv}_{S}$, $\lambda^{geo}_{S}$ are set at 0.01, 1, 5, 1, respectively. For semantic segmentation by MinkowskiNet, the maximum point number of each scan is 80,000 for SemanticKITTI and 50,000 for SemanticPOSS; The voxel size is 0.05 and we use coordinates and intensity of point cloud as input features.

\renewcommand\arraystretch{1.0}
\setlength{\tabcolsep}{0.9mm}{
\begin{table*}[t]
\centering
\begin{footnotesize}
\begin{tabular}{c|ccccccccccccccccccc|c}
 \toprule
  Method & \rotatebox{90}{car} & \rotatebox{90}{bi.cle} & \rotatebox{90}{mt.cle} & \rotatebox{90}{truck} & \rotatebox{90}{oth-v.} & \rotatebox{90}{pers.} & \rotatebox{90}{bi.clst} & \rotatebox{90}{mt.clst} & \rotatebox{90}{road} & \rotatebox{90}{parki.} & \rotatebox{90}{sidew.} & \rotatebox{90}{oth-g.} & \rotatebox{90}{build.} & \rotatebox{90}{fence} & \rotatebox{90}{veget.} & \rotatebox{90}{trunk} & \rotatebox{90}{terra.} & \rotatebox{90}{pole} & \rotatebox{90}{traf.} & mIoU\\
 \midrule
  Source-Only & 42.0 & 5.0 & 4.8 & 0.4 & 2.5 & 12.4 & 43.3 & 1.8 & 48.7 & 4.5 & 31.0 & 0.0 & 18.6 & 11.5 & 60.2 & 30.0 & 48.3 & 19.3 & 3.0 & 20.4 \\
  ADDA~\citep{tzeng2017adda} & 52.5 & 4.5 & 11.9 & 0.3 & 3.9 & 9.4 & 27.9 & 0.5 & 52.8 & 4.9 & 27.4 & 0.0 & 61.0 & 17.0 & 57.4 & 34.5 & 42.9 & 23.2 & 4.5 & 22.8 \\
  Ent-Min~\citep{vu2019advent} & 58.3 & 5.1 & 14.3 & 0.6 & 1.8 & 14.3 & 44.5 & 0.5 & 50.4 & 4.3 & 34.8 & 0.0 & 48.3 & 19.7 & 67.5 & 34.8 & 52.0 & 33.0 & 6.1 & 25.5 \\
  ST~\citep{zou2019CRST} & 62.0 & 5.0 & 12.4 & 1.3 & 9.2 & 16.7 & 44.2 & 0.4 & 53.0 & 2.5 & 28.4 & 0.0 & 57.1 & 18.7 & 69.8 & 35.0 & 48.7 & 32.5 & 6.9 & 26.5 \\
  \midrule
  PCT & 53.4 & 5.4 & 7.4 & 0.8 & 10.9 & 12.0 & 43.2 & 0.3 & 50.8 & 3.7 & 29.4 & 0.0 & 48.0 & 10.4 & 68.2 & 33.1 & 40.0 & 29.5 & 6.9 & 23.9 \\
  ST+PCT & 70.8 & 7.3 & 13.1 & 1.9 & 8.4 & 12.6 & 44.0 & 0.6 & 56.4 & 4.5 & 31.8 & 0.0 & 66.7 & 23.7 & 73.3 & 34.6 & 48.4 & 39.4 & 11.7 & 28.9 \\
 \bottomrule
\end{tabular}
\end{footnotesize}
\caption{
Experiments on unsupervised domain adaptation with SynLiDAR (as source) and SemanticKITTI (as target): PCT translates SynLiDAR and mitigates domain gaps in the input space effectively. It complements ST and combining them outperforms the baseline (\textit{i.e.}, source-only) by large margins. 
}
\label{tab:uda_kitti}
\end{table*}}

\renewcommand\arraystretch{1.0}
\setlength{\tabcolsep}{1.8mm}{
\begin{table*}[t]
\centering
\begin{footnotesize}
\begin{tabular}{c|ccccccccccccc|c}
 \toprule
  Method & pers. & rider & car & trunk & plants & traf. & pole & garb. & buil. & cone. & fence & bike & grou. & mIoU\\
 \midrule
  Source-Only & 3.7 & 25.1 & 12.0 & 10.8 & 53.4 & 0.0 & 19.4 & 12.9 & 49.1 & 3.1 & 20.3 & 0.0 & 59.6 & 20.1\\
  ADDA~\citep{tzeng2017adda} & 27.5 & 35.1 & 18.8 & 12.4 & 53.4 & 2.8 & 27.0 & 12.2 & 64.7 & 1.3 & 6.3 & 6.8 & 55.3 & 24.9\\
  Ent-Min~\citep{vu2019advent} & 24.2 & 32.2 & 21.4 & 18.9 & 61.0 & 2.5 & 36.3 & 8.3 & 56.7 & 3.1 & 5.3 & 4.8 & 57.1 & 25.5 \\
  ST~\citep{zou2019CRST} & 23.5 & 31.8 & 22.0 & 18.9 & 63.2 & 1.9 & 41.6 & 13.5 & 58.2 & 1.0 & 9.1 & 6.8 & 60.3 & 27.1 \\
  \midrule
  PCT & 13.0 & 35.4 & 13.7 & 10.2 & 53.1 & 1.4 & 23.8 & 12.7 & 52.9 & 0.8 & 13.7 & 1.1 & 66.2 & 22.9 \\
  ST + PCT & 28.9 & 34.8 & 27.8 & 18.6 & 63.7 & 4.9 & 41.0 & 16.6 & 64.1 & 1.6 & 12.1 & 6.6 & 63.9 & 29.6 \\
 \bottomrule
\end{tabular}
\end{footnotesize}
\caption{
Experiments on unsupervised domain adaptation with SynLiDAR (as source) and SemanticPOSS (as target): PCT translates SynLiDAR and mitigates domain gaps in the input space effectively. It complements ST and combining them outperforms the baseline (\textit{i.e.}, source-only) by large margins. 
}
\label{tab:uda_poss}
\end{table*}}

\subsection{Experiments on Data Augmentation}
We first evaluate how SynLiDAR augments real point cloud data as compared with state-of-the-art data augmentation methods as shown in Tables \ref{tab:da_kitti} and \ref{tab:da_poss}. We can observe that incorporating SynLiDAR helps train better models for both SemanticKITTI (+2.2\%) and SemanticPOSS (+2.6\%). It also shows that SynLiDAR shares good similarity with real-world dataset and provides a high-quality data source for transfer learning of LiDAR point cloud.

We also evaluate how PCT mitigates the domain gap and improves data augmentation. Experiments show that including PCT-translated SynLiDAR in training improves the mIoU by 2.2\% and 2.6\%, respectively, for SemanticKITTI and SemanticPOSS. The mIoU improvements clearly demonstrate the effectiveness of PCT in mitigating domain gap between SynLiDAR and the two real datasets.

\begin{figure}[t]
    \centering
    \includegraphics[width=0.9\linewidth]{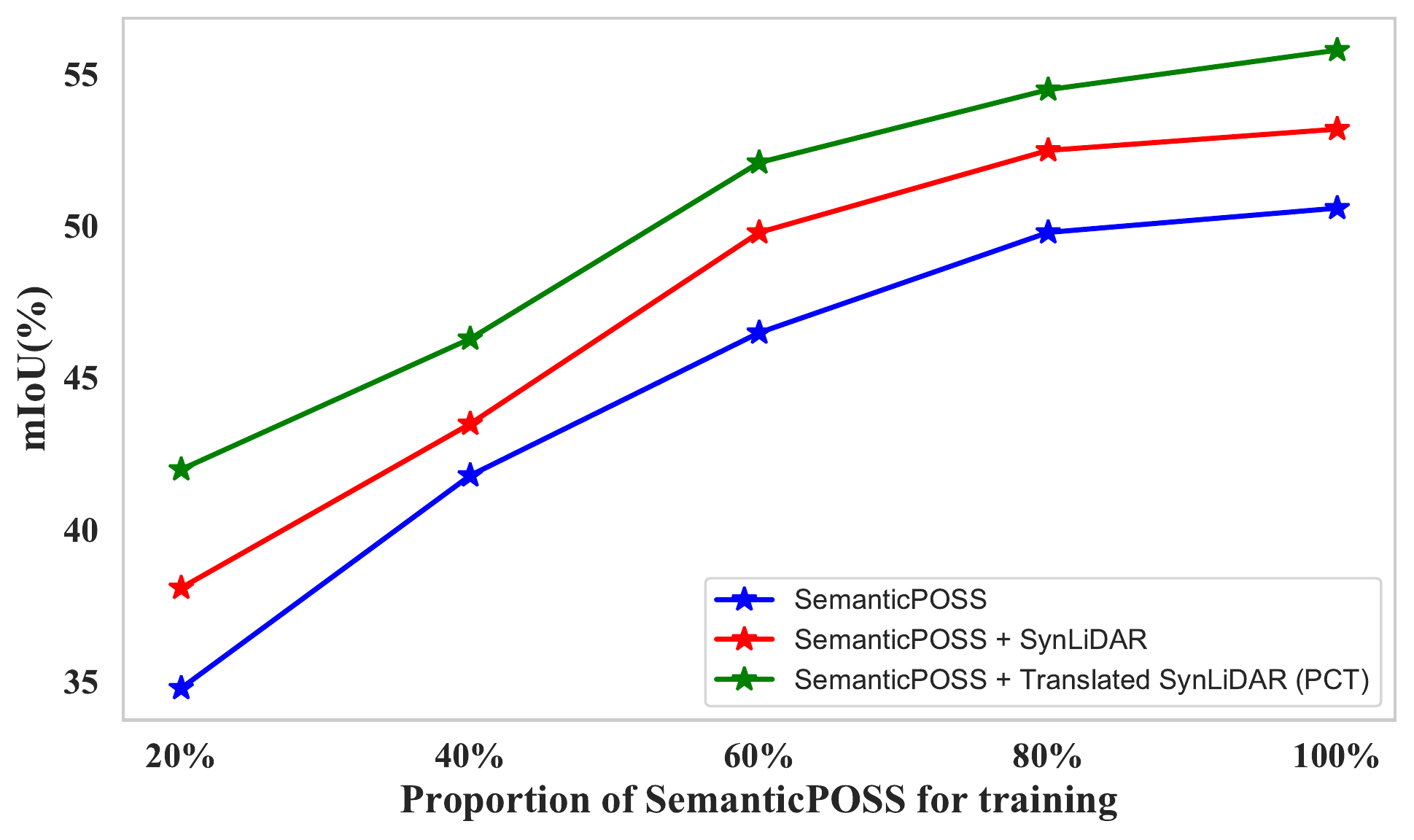}%
    \caption{
    SynLiDAR can effectively augment real-world LiDAR point cloud (SemanticPOSS) in a point cloud segmentation task. The PCT translated SynLiDAR further improves the augmentation consistently by large margins.
    }
    \label{fig:da_props}
\end{figure}

We further evaluate by combining SynLiDAR and PCT-translated SynLiDAR with different portions of SemanticPOSS, aiming to examine how SynLiDAR could alleviate data collection and annotation efforts. Fig. \ref{fig:da_props} shows experiment results. We can see that incorporating SynLiDAR consistently improves the segmentation under different portions of SemanticPOSS, and it can save up to 40\% SemanticPOSS without sacrificing segmentation performance. In addition, including the PCT-translated SynLiDAR further improves the segmentation consistently under similar setups. 

\subsection{Experiments on SSDA}
In this subsection we evaluate the effectiveness of PCT in another setup of semi-supervised domain adaptation (SSDA) with SynLiDAR (as source) and real datasets (as target).
We follow the setting of ~\cite{saito2019semi}
with three parts of training data, \textit{i.e.} labeled source samples, unlabeled target samples and 1 labeled target sample that are randomly selected for 1-shot SSDA-based semantic segmentation. 

Tables \ref{tab:ssda_kitti} and \ref{tab:ssda_poss} show experimental results of PCT and other state-of-the-art SSDA methods. As we can see, training labeled SynLiDAR and one-shot real sample with supervised loss (S+T) does not perform well for both two real datasets due to the domain gap.
Including PCT-translated SynLiDAR in training improved the mIoU by 2.2\% and 4.5\% for SemanticKITTI and SemanticPOSS, respectively.
Since PCT mitigates the domain gap in the input space while the state-of-the-art method APE~\cite{kim2020APE} does in the feature space, these two methods are complementary and combining them reached new state-of-the-art performances at 27.0\% and 31.2\%, respectively.

\subsection{Experiments on UDA}
In this subsection, we focus on unsupervised domain adaptation (UDA) for point cloud segmentation.
Different from SSDA, in this setup we only use labeled source data (SynLiDAR) and unlabeled target data (real datasets) for training.

As we can see from Tables \ref{tab:uda_kitti} and \ref{tab:uda_poss}, training on labeled source data as source-only failed to learn satisfactory segmentation models due to presence of the domain gap. State-of-the-art UDA methods mitigate the domain gap in either feature space (ADDA) or output space (Ent-Min, ST) and improved the segmentation performance effectively. On the other hand, PCT mitigates the domain gap in input space and including its translated SynLiDAR improved mIoU by 3.5\% and 2.8\% for SemanticKITTI and SemanticPOSS, respectively. It also complements ST and their combination achieved new state-of-the-art performances at 28.9\% and 29.6\%, respectively. 
The experiment results further indicate that PCT effectively reduced the domain gap between SynLiDAR and two real datasets.

\section{Conclusion}
This paper presents SynLiDAR and a point cloud translation method PCT for synthetic-to-real transfer learning. SynLiDAR is a large-scale synthetic LiDAR sequential point cloud dataset that contains 19 billion points with point-wise annotations of 32 semantic classes. PCT translates synthetic point clouds to have similar appearance and sparsity as real point clouds. 
Extensive experiments showed that SynLiDAR shares high geometrical similarities with real-world data,  
which effectively boosts semantic segmentation while combining with different proportions of real data.
PCT mitigates synthetic-to-real gaps effectively and its translated data further improves point cloud segmentation consistently in three transfer learning setups. 

\bibliography{aaai22}

\end{document}